\documentclass[12pt,letterpaper]{article}

\usepackage{cite}
\usepackage{textcomp}
\usepackage{xcolor}
\usepackage{url}

\usepackage{geometry}
\usepackage[compact]{titlesec}

\usepackage{caption}
\usepackage{enumitem} 
\usepackage[font+=footnotesize]{subcaption}
\usepackage{amsmath,epsfig,amsfonts,amssymb,graphicx,bm,amsbsy,amssymb,url,tabularx,latexsym}
\usepackage{graphicx}
\usepackage{verbatim}
\usepackage{algorithm, algorithmic}

\newcommand{\xspace}[0]{{\cal X}}

\newcommand{\Mspace}[0]{{\cal M}}
\newcommand{\sspace}[0]{{\cal S}}
\newcommand{\ispace}[0]{{\cal I}}

\newcommand{\moreone}[0]{[1,\infty]}

\newtheorem{theorem}{Theorem}
\newtheorem{lemma}{Lemma}

\newenvironment{proofnumbered}{\noindent{\em Proof}}{\hfill $\square$}

\DeclareMathOperator*{\argmin}{argmin}
\DeclareMathOperator*{\argmax}{argmax}

\def\BibTeX{{\rm B\kern-.05em{\sc i\kern-.025em b}\kern-.08em
    T\kern-.1667em\lower.7ex\hbox{E}\kern-.125emX}}

\begin{document}

\title{Optimal Scheduling in  a Question-Answering Forum of Knowledge Workers 
}

\author{Rohit Negi and Mustafa Yilmaz  \\
	Carnegie Mellon University  \\
	}
\date{}

\maketitle

\begin{abstract}
As individuals turn to the Internet to find answers to questions they may have, several Question Answering (QA) forums have evolved, where users knowledgeable in certain topics can contribute their expertise to answering these requests for information. While these are currently  volunteer based, we consider a future version employing knowledge workers who are experts in certain topics. In such a system,  the request-answer processes forming the  queuing system may utilize  schedulers that  assign requests in different topics to the experts in the  forum, who may be able to answer them according to their expertise levels in different topics. With this model, we calculate the capacity of the system for handling the requests while keeping the system stable, and design  schedulers that achieve  capacity. We also investigate how collaboration between experts in answering requests can potentially increase capacity.
\end{abstract}

{\bf Keywords}:
question answering, social networks, scheduling.

\vspace{6truemm}

\section{Introduction}
Question Answering (QA) forums \cite{cqa}  such as {\tt Quora} and {\tt StackExchange}  \cite{socnetworkexamples} allow individuals to ask questions, so that users in the  QA forum, knowledgeable in that topic, can answer these based on their own knowledge, and perhaps augmented with search for relevant information using information sources. Posted questions may contain  keywords (tags) or may be posted in a specific narrow forum within the QA system, so that the topic of the question is clear to potential answerers knowledgeable in that topic.  QA forums have sparked a variety of research, such as automatic identification of topics based on textual information \cite{nassif},   diffusion of information in online social networks  \cite{dhamal} and models to understand effects of such diffusion on human behavior  \cite{fan}.

While the present volunteer driven QA forums are certainly successful, a certain fraction of questions in them remains unanswered, due to lack of volunteer interest or expertise in answering those questions. For example, analysis of retrieved archival data \cite{internetarchive} shows that 16\% of questions remain unanswered and 54\% of questions do not have an answer marked `Accepted' out of the 1.8 million questions on  37 most popular  {\tt StackExchange} websites on technical topics.
Similarly, certain questions receive active interest from volunteers, with several dozen answers, while other questions languish without volunteer interest. At the same time, the rise of the large language models, such as the Chatgpt bot currently being  tested by the QA forum {\tt Quora}, may put a premium on {\em dedicated} human experts answering  complex questions that bots cannot easily answer. Thus, while researchers are investigating  incentives for volunteers in crowdsourcing applications to complete tasks such as question-answering  \cite{huang}, it may be worth imagining a future version of these QA forums (or perhaps a premium tier within an existing QA forum), which, by monetizing the forum, is able to employ dedicated knowledge workers, whom we will call {\em experts}, to guarantee an acceptable level of performance in terms of complex questions being successfully answered.  Such guarantees may be obtained by  requiring these knowledge workers to be available as contracted, as well as dynamically assigning specific questions to these workers, using a scheduler, with a view to
 increasing the total system capacity.  Even if such as knowledge worker-based forum (or a premium tier) doesn't come to fruition, a theoretical analysis of the proposed system will also provide an upper bound on the capacity of a volunteer driven QA forum, since volunteers are not necessarily motivated by the express goal of maximizing system capacity.
 
 This paper investigates the problem of designing such a scheduler for the proposed knowledge worker-based system, to maximize request answering capacity without the experts becoming overloaded. We  propose a system model where requests for information on different topics arrive and are placed in topic queues. In previous research \cite{negi2022} based on this model, we had proposed a fixed scheduler that assigns different fractions of requests from the topic queues to different experts. However, that scheduler required {\em offline calculations} of the fractions assigned, based on complete knowledge of the statistics of the request arrival  processes, as well the capabilities of all the experts. In this paper, we propose an {\em online  scheduler} that assigns each request dynamically from the queues to  experts based on the latter's expertise in different topics and the current state of the system, so that the question is researched by one or more experts until it is answered. Further, we investigate a {\em coordinated} mode of operation, where different experts handle different requests, and also a {\em collaborative} mode, where several experts jointly work on a request, by pooling together their expertise. Queuing theoretic analysis of the scheduler is used to calculate the capacity  of the QA system. Insights into the capacity are obtained and  the benefit of collaboration between the experts versus mere coordination is explored. While there is vast prior work on analysis of scheduling  \cite{meyn} in diverse applications, which we can draw on for analysis, the focus of this paper on QA forums reveals interesting insights into the capacity of  knowledge-handling systems such as these forums.

\section{Model}
\label{sec:model}

We assume that individuals can ask {\em questions}, also called {\em requests} for information, in the QA forum. Each question belongs to a topic $x \in \xspace$, with $\xspace$ being a {\em finite} set of possible topics, and with the questions' topic being identified using keywords marked in the question,  by an automatic classifier based on the question's text, or simply posted to a narrow sub-forum that specializes in that topic. We assume that time is slotted $t=1,2,\ldots$, and a random number $a_x(t)$ questions on topic $x$ arrive at the end of slot $t$, where $a_x(t)$ are i.i.d. in time, and independent across $x$. We assume that $E[a_x(t)] =\lambda p(x)$ with the probability mass function (p.m.f.) $p(x)>0, \forall x$ representing the relative frequency of different topics, and request load $\lambda$ being the mean number of questions in each slot. We assume that the variance of $a_x(t)$ is bounded by $c_a \lambda^2$ for some constant $c_a$. Topic queues, one for each topic, are maintained for unanswered questions, with $Q_x(t)$ being the queue length for topic $x$ at the beginning of time $t$. 

There are $n$ available dedicated experts denoted $i=1,2,\ldots,n$ that can monitor the QA forum. We assume that the experts agree to use a scheduler to assign questions to them, so that these experts are {\em coordinating} with each other. (Thus, the model is explicitly {\em not} considering volunteer-based forums, where the motivations of the volunteers may not align with any notion of system capacity.) Call $\sigma_{x,i}(t)=1,0$ according to as expert $i$ is assigned a request from topic $x$ at time $t$, or not, respectively. We assume that an expert $i$ can work on (or {\em research}) no more than $n_i$ requests at each $t$, where experts who can dedicate more time may choose a higher value of $n_i$, similar to how it works for drivers in the ride-share industry. Expert $i$ succeeds in answering a question in a given slot with a probability of $q_i(x)$ (which could be zero), which we call the {\em answering rate} and depends on the topic $x$. We call $d_{x,i}(t)$ the number of successful answers provided by expert $i$ in topic $x$ at time $t$. If the expert is unsuccessful, the request goes back to its topic queue, and may be assigned to another (or the same) expert in subsequent time slots. (We experimentally show that this assumption does not seem to be not very critical to the results - see Section \ref{sec:simulation}. We also  design a separate scheduler that explicitly allows workers to
work on a request until completion.) 

We assume that the probability of successfully answering that request does not depend on its past history, so that the number of (possibly non-consecutive) time slots $\mbox{T}$ needed by expert $i$ to answer the request is a geometric random variable with values $1,2,\ldots$ and mean value $T_i(x) = \frac{1}{q_i(x)} \geq 1$. (While this is a useful assumption for technical reasons, we show experimentally in Section \ref{sec:simulation} that this assumption  seems to be adequate from a practical perspective, provided an upper bound is placed on $\mbox{T}$.) Random variable $\mbox{T}$ is apriori unknown to the scheduler, although we assume that its mean value, the average research time $T_i(x)$, is known for all $i,x$.  Thus, as far as scheduling is concerned, an expert $i$ is simply a function $T_i:\xspace \rightarrow \moreone$. The random answering time accounts for varying difficulty in searching for information, or processing that information, for different questions, even in the same topic. It also accounts for the back and forth refining of questions, based on partial answers, that happens in QA forums. The $T_i(\cdot)$ functions will generally be different for different experts to account for their specific expertise.  The simplest method to estimate $T_i(x)$ is to use past history of the expert's answers in different topics. More sophisticated methods, such as in  \cite{han}, may also potentially be used. We assume that the scheduler can use past and current history, including queue lengths $Q_x(t)$, to decide on $\sigma_{x,i}(t), \forall x,i$. Since we assumed that arrivals occur at the end of a slot,  the queue lengths update as
\begin{eqnarray}
Q_{x}(t+1)=Q_{x}(t)+a_{x}(t)-\sum_{i=1}^n d_{x,i}(t), \ \forall x  \label{eqn:queueupdate} 
\end{eqnarray}

We call the QA system as unstable if, for any scheduler that uses past and current history, diverging queue lengths are possible, i.e.,
\begin{eqnarray}
P(\lim_{t \rightarrow \infty} \sum_{x \in \xspace} Q_{x}(t) = \infty) > 0. \label{eqn:instability}
\end{eqnarray}
To prove stability, we will only consider schedulers that use current queue lengths $Q_x(t)$ at time $t$. For such schedulers, due to the i.i.d. arrivals, and the geometric distribution of answering times, the QA system is a Markov chain with  $Q_x(t), \forall x$ being the state at time $t$. We say the system
is stable if this Markov chain is positive recurrent \cite{bremaud}. In particular, this would also mean that the system is not unstable in the sense of \eqref{eqn:instability}. We say $\lambda$ is achievable if a scheduler can keep the system stable. The supremum of all such $\lambda$ is the {\em capacity} of the system. 

For conciseness, we generally omit writing the set of an index, when that set is obvious, e.g., writing $\sum_x$ instead of  $\sum_{x \in \xspace}$. Similarly, we will write $\max_{\alpha_i}$ instead of $\max_{\alpha_i,i=1,2,\ldots,n}$.
 Due to lack of space, we leave the proofs to a subsequent full length paper, while noting that they rely on Lyapunov analysis, but for the special types of graph theoretic problems encountered in this formulation. While the capacities we obtain can be conceivably used to design or evolve a QA forum  (such as by hiring experts that can increase the computed capacity), we also  provide an intuitive understanding of the capacity expressions.

\section{Coordination Capacity}
\label{sec:multipleonline}

Suppose that the $n$ experts decide to coordinate as described in Section \ref{sec:model}, by using a scheduler that decides the assignment of unanswered questions to each expert in each slot $t$, with a view to maximizing the achievable request load $\lambda$.  Let $\Mspace_i \doteq \{x: q_i(x)>0\}$ be the set of topics that expert $i$ has some capability of answering. We assume that $\Mspace_i \ne \emptyset, \forall i$, since experts that can answer no topics don't need to be considered for scheduling.  
In this case, we have the following result.

\begin{lemma}
\label{thm:multiple}
The capacity with  coordinating experts  is $\lambda^* = 0$ if there exists a topic $x$ having $\max_i q_i(x)=0$;
otherwise it is strictly positive and given as
\begin{eqnarray}
\lambda^* & = & \left(\max_{\alpha_i} \sum_{x \in \xspace}  \ \min_{i: \, q_i(x)>0} \left( \alpha_i \frac{p(x)}{q_i(x)} \right) \right)^{-1} \ \mbox{where} \label{eqn:multiplecoordinating} \\
&&  \sum_{i=1}^{n} \alpha_i n_i =1, \quad \alpha_i  \geq 0, \ \forall \, i=1,\ldots,n. \label{eqn:multiplecoordinatingalpha}
\end{eqnarray}
Further, any $\lambda < \lambda^*$
can be achieved using a queue length based scheduler, such as the  Greedy Online Coordinating scheduler described below. 
\end{lemma}

{\bf Greedy Online Coordinating  scheduler}: 
 In each time slot $t$, for each expert $i$ {\em separately}, the scheduler first sorts the requests in the topic queues $x$ in decreasing order of rewards $r_i(x) = Q_{x}(t) q_i(x)$. It then tentatively assigns $n_i$ of the largest reward requests to expert $i$. Then, in the second stage, it finalizes the requests by arbitrarily deleting  excess tentative requests in each topic $x$ that exceed the number $Q_x(t)$ waiting in that topic queue, and presents them to  the experts. 
 
 Note that this is an Online 
algorithm since the scheduler here does not need to pre-compute fixed fractions of requests to be
assigned to different experts, unlike the case in \cite{negi2022}, which would require knowledge of the arrival $p(x)$, and can potentially also adapt here if this p.m.f. changes slowly.
The scheduler can be implemented in a {\em federated} manner, as follows. Each expert can independently compute its own rewards $r_i(x)$ and choose the largest $n_i$ of these to announce its tentative schedule to the scheduler. The scheduler can then run the second phase of the greedy assignment after collecting all these tentative schedules. Thus, the values of $q_i(x)$ only needs be known by the corresponding expert $i$.

\begin{proofnumbered}
{\em [Lemma \ref{thm:multiple}]:} Coordination is a special case of Collaboration, as discussed in Section \ref{sec:multiplecollaborate}, so that this theorem is a special case of
Theorem \ref{thm:collaborating}, obtained by setting $\sspace=\{\{i\},i=1,2,\ldots,n\}$,
the set of all singletons. Further the Greedy scheduler proposed above is a special case of
the Two-phase Greedy scheduler described in Algorithm \ref{alg:greedytwophase} for the Collaboration case. Although in the Collaboration case, the Two-phase Greedy algorithm can only guarantee an
approximation ratio of $\gamma \doteq (\max_{S \in \sspace} |S|)^{-1}$ with respect to capacity (Lemma \ref{thm:collaboratingapproximatescheduler}),
in the Coordination case, $\gamma=1$ since the set $\sspace$ only consist of singleton sets $\{i\}$. Thus,
in this case, the Greedy scheduler achieves capacity.
This Lemma generalizes the one obtained in \cite{negi2022} for {\em offline} scheduling.
\end{proofnumbered}

Coordinating experts can potentially compensate for each others' lack of expertise in certain topics. We calculate the coordination capacity  in certain special cases to provide intuition on that aspect. To avoid clutter here, we assume $n_i\equiv 1$.

\begin{lemma}
\label{thm:multiplespecialcases}
The coordination capacity $\lambda^*$  in Lemma \ref{thm:multiple} is bounded as follows. 
\begin{enumerate}[label=(\alph*)]
    \item $\lambda^* \leq n$ always.
    \item {\em Competency of experts}: Suppose for each topic $x$, at least $N$ experts have answering rate $q_i(x) \geq q$. Then, $\lambda^* \geq qN$.
    \item {\em Diversity of experts}: Suppose a partition of topics $S^{(i)},i=1,\ldots,n$ exists (allowing $S^{(i)}=\emptyset$), with each $S^{(i)} \doteq \{x: q_i(x)\geq q\}$. Let $p \doteq \max_i P(X \in S^{(i)})$. Then, $\lambda^* \geq \frac{q}{p}$, but it requires $n \geq \frac{1}{p}$ experts.
\end{enumerate}
\end{lemma}

Part b) states that if every topic has several competent experts, capacity is proportionally higher. 
Practically, we expect an expert to be competent in only a subset of topics. If many competent experts exist, but they only cover similar topics, requests in other topics will not be sufficiently answered. Part (c) states that capacity will be high if many experts exist, each competent in a subset of topics, such that all topics are covered. In part (c), $P(X \in S) = \sum_{x \in S} p(x)$. By partition, we mean that $S^{(i)} \cap S^{(j)} = \emptyset, \forall i \ne j$ and $\cup_{i} S^{(i)} = \xspace$.

 In QA forums like {\tt Quora} and {\tt StackExchange}, 
multiple experts may propose answers to a question until that question is resolved. At the same time, an expert may work on other requests before
coming back to a previous request she had worked on. 
The above Online Coordinating  scheduler models that behavior. If an expert is unsuccessful in finding the answer, the request goes back to its queue, and may subsequently be worked
on by a different  (or same) expert, and each expert is allowed to work on a different request in subsequent time slots. However,
in other QA forums such as Microsoft Community \cite{socnetworkexamples},  each request is often handled by only a single moderator or `MVP', until it gets answered. While experiments in Section \ref{sec:simulation} suggest that
the above scheduler potentially remains optimal even if the model is changed to allow an expert
to keep working on a request until she answers it, the following scheduler explicitly models such non-preemptive behavior.
\par
{\bf Online Arrival scheduler}: This scheduler maintains separate topic queues $Q_{x,i}(t)$ for each expert $i$. Choose an arbitrary
$c_Q>0$.
In each time slot $t$, each expert $i$ chooses $n_i$ requests arbitrarily from among those queued up at her queues $Q_{x,i}$, or continues working on the requests from the previous slot.
At the end of that time slot,  the scheduler assigns all  $a_x(t)$ arriving requests in topic $x$ to the corresponding topic queue of a specific expert $i^*(x)$ 
chosen as below.
\begin{eqnarray}
i^*(x) \  =  \ \argmin_{i: \, q_i(x)>0} \  \frac{1}{q_i(x)} (c_Q+ \sum_{x\in \Mspace_i} \frac{1}{n_i q_i(x)}Q_{x,i}(t)).  \label{eqn:multiplecoordinatingscheduleronlinearriving} 
\end{eqnarray}
In the argmin above, if there is a topic $x$ for which $q_i(x)=0, \forall i$, those requests are discarded. (By Lemma \ref{thm:multiple}, the capacity is zero in this degenerate case.)
Only a single expert  works on each given  request until it
gets answered.
The Lemma below shows that this scheduler is also optimal.
\begin{lemma}
\label{thm:multipleonlinearriving}
The Online Arrival scheduler \eqref{eqn:multiplecoordinatingscheduleronlinearriving} for  coordinating experts   achieves the same capacity as shown in Lemma \ref{thm:multiple}.
\end{lemma}

 In this scheduler, the constant $c_Q$ is
required because queue lengths are natural numbers, and so, an empty queue conveys no information about lack of expertise in some topic, unless $c_Q>0$.  Simulations in Section \ref{sec:simulation} do not indicate a strong effect of $c_Q$, with  $c_Q=1$ being adequate.
The Online Arrival scheduler can also be implemented in a federated manner, similar to the two-phase Greedy scheduler.

\section{Collaboration Capacity}
\label{sec:multiplecollaborate}

In this section, we extend the model of Section \ref{sec:model}, by assuming that that certain sets
of experts can {\em collaborate}.  This means
that they can jointly research a common request $x$, sharing their diverse expertise, brain-storming or identifying strengths and weaknesses of each others' arguments, and then jointly arrive at the correct answer after a random amount of time, modeled as a positive geometric random variable.

To model collaboration, denote by $q_S(x)$  the probability of answering a request $x$ in a given time slot
if experts in the set
$S$ collaborate on $x$. (So, $T_S(x)=1/q_S(x)$ is the average time
to collaboratively answer the request.) For example,
the {\em collaboration set} $S=\{1,2,5\}$ if experts 1,2,5 collaborate.  Let 
$\sspace$ be the set of all feasible collaborations. $\sspace$ is assumed to always contain all the singletons $\{i\}, i=1,2,\ldots,n$,
which corresponds to the  non-collaborative research mode of individual experts. For these singletons, $q_{\{i\}}\doteq q_i(x)$ is the usual
non-collaborative answering rate. $\sspace$ need not be a power set of the experts, since not all
experts may want to collaborate with each other, and further, collaborations may practically be limited to only a
few experts at a time. We assume that an expert $i$ can
participate in at most $n_i$ collaboration sets $S \in \sspace$ in a given time slot, since that is
the number of problems he may simultaneously work on. Thus, if $I \subset \sspace$ is the set of collaboration sets that are
assigned requests by the scheduler in a given time slot, expert $i$ cannot appear in more than $n_i$ sets in $I$. Note that for the special case $n_i\equiv 1$, $I$ must be an independent set.  i.e., $S_1,S_2 \in I \Rightarrow |S_1 \cap S_2| = \emptyset$. So, we can call the $n_i$-constrained version as a `generalized independent set'. Let $\ispace$ be the set of {\em maximal} generalized independent sets. i.e., if a generalized independent set $I$ is in $ \ispace$, then 
 there is no other generalized independent set $\tilde{I}$ in $ \ispace$ containing $I$. For example, for $n_i\equiv 1$, if the only collaboration sets are $a=\{1\}, b=\{2\},c=\{3\}, d=\{1,2\}$, then $\ispace$ consists of  $\{a,b,c\}$ and $\{c,d\}$ only.

Let $\Mspace_S \doteq \{x: q_S(x)>0\}$ be the set of topics that collaboration set $S$ has some capability of answering. For collaborative research, we have the following capacity result.
\begin{theorem}
\label{thm:collaborating}
The capacity with collaborative research is $\lambda^* = 0$ if there exists a topic $x$ having $\max_S q_S(x)=0$; otherwise it is strictly positive and given as
\begin{eqnarray}
\lambda^* \ = \  \left(\max_{\alpha_S} \sum_{x \in \xspace} \ \min_{S:\,  q_S(x)>0} \left( \alpha_S \frac{p(x)}{q_S(x)} \right) \right)^{-1} \  \mbox{where} \label{eqn:collaborating} \\
 \sum_{S \in I} \alpha_S \leq 1, \ \forall \, I \in \ispace, \qquad \alpha_S \geq 0,  \ \forall S \in \sspace.  \label{eqn:alphaSconstraint}
\end{eqnarray}
Further, any $\lambda < \lambda^*$
can be achieved using a queue length based scheduler, such as the one shown below. 
\end{theorem}

\begin{algorithm}[t]
\caption{: Two-phase Greedy Collaboration scheduler}
\label{alg:greedytwophase}
\begin{algorithmic}
\STATE $\sspace_0=\sspace, \ \ U=\emptyset,  \ \ n(x)\equiv 0, \ \ \sigma_{x,S}(t)\equiv 0,\ \  N_i\equiv 0$;
\WHILE{($\sspace_0 \ne \emptyset$)}
\STATE $x^*,S^* = \argmax_{x,\, S \in \sspace_0} \, Q_x(t)q_S(x)$;
\STATE $\sigma_{x^*,S^*}(t)=1$; \ \ $n(x^*) \leftarrow n(x^*)+1$;
\STATE $N_i \leftarrow N_i+1, \ \forall i \in S^*$; \ \ $U \leftarrow U \cup \{i: N_i=n_i\}$;
\STATE $\sspace_0 \leftarrow \sspace_0 \setminus \{S: S \cap U \ne \emptyset \}$;
\ENDWHILE
\FOR{$x \in \xspace$}
\IF{$n(x) > Q_x(t)$}
\STATE   Arbitrarily delete $n(x)-Q_x(t)$  assignments of topic $x$ requests made to collaboration sets;
\ENDIF
\ENDFOR
\end{algorithmic}
\end{algorithm}

{\bf Online Collaborative scheduler}: In each time slot $t$,  the scheduler assigns requests from topic queues
$x$ to 
the collaboration sets $S$ (denoted $\sigma_{x,S}(t)=1$, else $0$),
by solving the  Hypergraph Assignment Problem (HAP) below. \cite{hap}. 
\begin{eqnarray}
R^*  & =&  \max_{\sigma_{x,S}} \  \sum_x Q_{x}(t) \sum_{S \in \sspace}  q_S(x)\sigma_{x,S}(t) \quad \mbox{where} \label{eqn:collaborativescheduleronline} \\
& & \mkern-36mu  \sum_x \sum_{S: \, i \in S} \sigma_{x,S}(t)\leq n_i, \ \forall \, i=1,2,\ldots,n,  \label{eqn:expertconstraintcollaborative} \\
& & \mkern-36mu   \sum_{S \in \sspace} \sigma_{x,S}(t)\leq \min\{Q_{x}(t),n\}, \ \forall \, x \in \xspace.  \label{eqn:topicconstraintcollaborative}
\end{eqnarray}
Here, \eqref{eqn:expertconstraintcollaborative}
ensures that an expert $i$ is not assigned more than $n_i$ requests in a time slot, whether individually or
collaboratively, while \eqref{eqn:topicconstraintcollaborative} ensures that only requests
actually present in the topic queues are assigned.  The HAP  \eqref{eqn:collaborativescheduleronline} can be visualized as a type of bipartite graph matching problem to match requests with experts, but with constraints on which collaboration sets may be simultaneously active (due to the generalized independent set constraints.)

To demonstrate the possible advantage of collaboration, consider a stylized case where there are $n=2$  experts and they can collaborate.
Assume $n_i \equiv 1$ and that  $p(x_1)=1$, so that $x_1$ is the only topic of interest. Assume $q_1(x_1)=q_2(x_1)=\frac{1}{100}$ because each expert finds the request difficult to answer, 
while $q_{\{1,2\}}(x_1)=1 \gg q_1(x_1)+q_2(x_1)$ because
 the two experts can pool their complementary expertise to solve $x_1$ quickly.
Then by symmetry,  the solution to \eqref{eqn:collaborating} has  optimal $\alpha_{1}=\alpha_2=0.5$ and $\alpha_{\{1,2\}}=1$, so that $\lambda^* = 
\left(\sum_x \min\left(\alpha_{\{1,2\}} \frac{p(x)}{q_{\{1,2\}}(x)}, \ \min_i \left(\alpha_i \frac{p(x)}{q_i(x)}\right) \right) \right)^{-1} = \left(\min(1,50,50)\right)^{-1}=1$. 	On the other hand, the coordinating but
non-collaborating case \eqref{eqn:multiplecoordinating} yields with the optimal $\alpha_{1}=\alpha_2=0.5$, the capacity $\lambda^* = \left(\min(50,50)\right)^{-1}=0.02$, which is
significantly smaller than the capacity with collaboration. 

The extent of the advantage of collaboration over coordination depends strongly on the specific topic distribution $p(x)$ and
answering rates $q_S(x)$. The result below provides a comparison of the collaboration capacity $\lambda^*_{collab}$ in Theorem \ref{thm:collaborating} to the (non-collaborative) coordination capacity $\lambda^*_{coord}$ in Lemma \ref{thm:multiple}, under
a condition on answering rates. 
Call $\lfloor \cdot \rfloor$ as the floor function. 
 Let $\sspace^{(2+)}  \subset \sspace$ be the subset of non-singleton collaboration sets (i.e., having at least two experts.) Let $\ispace^{(2+)} \subset \ispace$ be the set of all maximal independent sets that consist of only sets in $\sspace^{(2+)}$. Let $\ispace_{cover} \subseteq \ispace^{(2+)}$ be a cover for all the experts, i.e., we  have $i \in \cup_{I \in \ispace_{cover}} \cup_{S \in I} S$ for every expert $i$. Such a cover  always exists if (and only if) every expert $i$ participates in some non-singleton collaboration set, i.e., every $i \in S$ for some $S \in \sspace^{(2+)} $. If no such $\ispace_{cover}$ exists, we can consider $|\ispace_{cover}|=\infty$ in the bound below. 

\begin{lemma}
\label{thm:collabcoordcomparison}
For $n_i \equiv 1$, the collaboration capacity $\lambda^*_{collab}$  in Theorem \ref{thm:collaborating} is bounded as follows. 
\begin{enumerate}[label=(\alph*)]
    \item $\lambda^*_{coord} \leq \lambda^*_{collab} \leq n$ always.
    \item Suppose that we have the uniform bound $q_S(x) \geq c \left(\max_{i \in S} q_i(x) \right), \forall x,S \in \sspace^{(2+)} $, where $c \geq 1$ is a constant. Then, $\lambda^*_{collab} \geq \left(\frac{c}{ |\ispace_{cover}| \, (\max_S |S|)}\right) \lambda^*_{coord}$.
    \item With the same assumptions as in part (b), if in addition, all possible collaborations of size $K$ are allowed (with $K \geq 2$), we have $\lambda^*_{collab} \geq \left( \frac{c}{K} \frac{\lfloor n/K \rfloor}{n/K} \right) \lambda^*_{coord}$.
\end{enumerate}
\end{lemma}

Note that the  bound on $q_S(x)$ assumed in part (b) is trivially true for $c=1$, since each
collaboration set $S$ can simply designate its best expert on topic $x$ to answer requests in that topic. The lemma shows that if collaboration always results in significantly faster answers, perhaps because experts can pool their expertise, or perhaps because experts function better when they `bounce ideas' off each other, the collaboration capacity is proportionally higher. In part (b), the reduction due to the collaboration set size $\max_S |S|$ is reasonable, since experts working together in $S$ on a single request lose the ability to work in parallel on several individual requests in that slot. The reduction due to $|\ispace_{cover}|$ accounts for many experts requiring the same few experts for collaboration, so that the latter get `spread too thin'.  In fact, this capacity bound is tight, as shown by
the two-expert example in the previous paragraph, for which we had $c=100, |S|=2$, $\ispace_{cover}=\{I\}$ with $I=\{\{1,2\}\}$, while $\frac{\lambda^*_{collab}}{\lambda^*_{coord}} = \frac{1}{0.02} =  \frac{c}{|\ispace_{cover}| \, \max_S |S|}$ there. In part (b), the bound also applies if only collaboration sets of size $K \leq \max_S |S|$ are considered, with appropriate change in the definition of  $\sspace^{(2+)}, \ispace^{(2+)}$. The symmetric case in part (c) shows a capacity
gain factor of approximately $\frac{c}{K}$.

Unfortunately, from a practical perspective, the Hypergraph Assignment Problem is known to be NP-hard \cite{hap}. So, for large $n,|\xspace|$,
approximate solutions to the HAP \eqref{eqn:collaborativescheduleronline} are of interest. Another practical scenario of interest is when experts have an erroneous estimate $\hat{T}_S(x)$ of the true average research times $T_S(x)$, since these may have been estimated by them using their past 
experience answering requests about different topics. Call $\hat{q}_S(x) = \frac{1}{\hat{T}_S(x)}$. 
The lemma below guarantees QA system stability in such cases, under certain assumptions.

\begin{lemma}
\label{thm:collaboratingapproximate}
Using the erroneous $\hat{q}_S(x)$ instead of $q_S(x)$ in both \eqref{eqn:collaborating} and \eqref{eqn:collaborativescheduleronline}, let $\lambda^*$ be the capacity computed using
\eqref{eqn:collaborating} and 
$R^*$ be the optimal reward achieved in HAP \eqref{eqn:collaborativescheduleronline}.
Suppose for fixed constant $\tilde{\gamma} \leq 1$, we have the uniform bound $\hat{T}_S(x) \geq \tilde{\gamma} T_S(x), \forall x,S$. Suppose also, for fixed  constants $\gamma \leq 1$, $c\geq 0$,  the HAP \eqref{eqn:collaborativescheduleronline} (with erroneous $\hat{q}_S(x)$)
can be solved approximately to obtain a reward $R$ bounded as $R \geq \gamma R^* - c$, where $R^*$ is the erroneous optimal reward.
Then, the  Collaborative scheduler \eqref{eqn:collaborativescheduleronline} can achieve any $\lambda < \tilde{\gamma} \gamma \lambda^*$.
\end{lemma}

Note that the true $q_S(x)$ need not be known to use the above lemma - only a guaranteed $\tilde{\gamma}$.
For $\gamma=1,c=0$, this reduces to the case of exactly solving the HAP, but with erroneous answering rates. If in addition, $\tilde{\gamma} = 1$, this reduces to Theorem \ref{thm:collaborating}.  This lemma also extends Lemma \ref{thm:multiple} to the case of erroneous answering rate,
since coordination is a special case of collaboration consisting of only the singleton collaboration sets.
An example of  an approximate scheduler to which Lemma \ref{thm:collaboratingapproximate} is applicable is described
below.

{\bf Two-phase Greedy Collaboration scheduler}:
In each time slot $t$,  the scheduler runs two phases to determine the assignments of requests to collaboration sets (Algorithm \ref{alg:greedytwophase}). $N_i$ is the number of
times expert $i$ appears in the collaboration sets already chosen until that point.  $n(x)$ counts the number of assignments of requests in topic $x$ that have already been made until this point in the slot. In the first phase, it repeatedly assigns $x$ to a collaboration set $S \in \sspace_0$ if $x,S$ maximizes $Q_x(t)q_S(x)$, while removing from $\sspace_0$ all sets that depend on experts $i$ that have used up their allotted $n_i$ assignments (set $U$.)  In the second phase, it deletes the surplus assignments that have been made beyond that topic's queue length.

\begin{lemma}
\label{thm:collaboratingapproximatescheduler}
The Two-phase Greedy scheduler solves the HAP \eqref{eqn:collaborativescheduleronline}
with a reward of at least $\gamma R^*-n^2$, with
 $\gamma \doteq (\max_{S \in \sspace} |S|)^{-1}$. Thus,
 it achieves 
 $\lambda < \gamma \lambda^*$ by Lemma
 \ref{thm:collaboratingapproximate}.
\end{lemma}

Note that in the above definition, $\gamma^{-1}$ is the maximum number of experts that
may collaborate on a given request.
For example, if only pairwise collaborations are
allowed, then $\gamma^{-1} = 2$, and so, the above greedy scheduler
will achieve at least half the capacity \eqref{eqn:collaborating} according to this lemma.

\begin{figure}[t]
\centering
\includegraphics[width=0.9\textwidth]{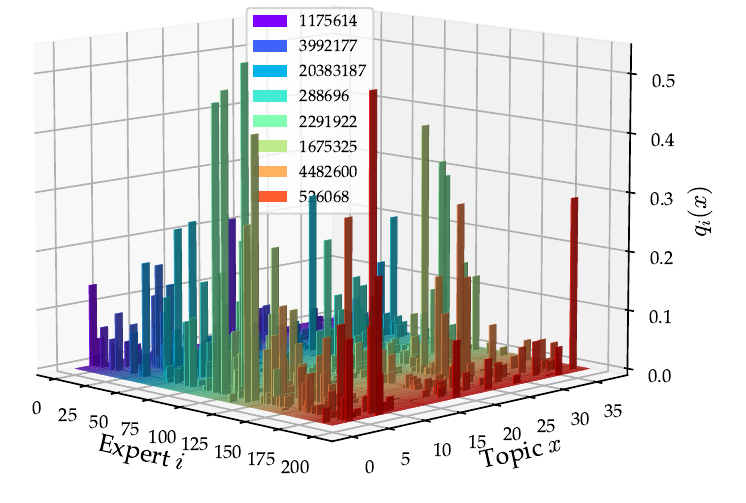}
\caption{$q_i(x)$ for some active users of {\tt StackExchange}.}
\label{fig:qix}
\end{figure}

\begin{figure}[t]
\centering
\includegraphics[width=0.9\textwidth]{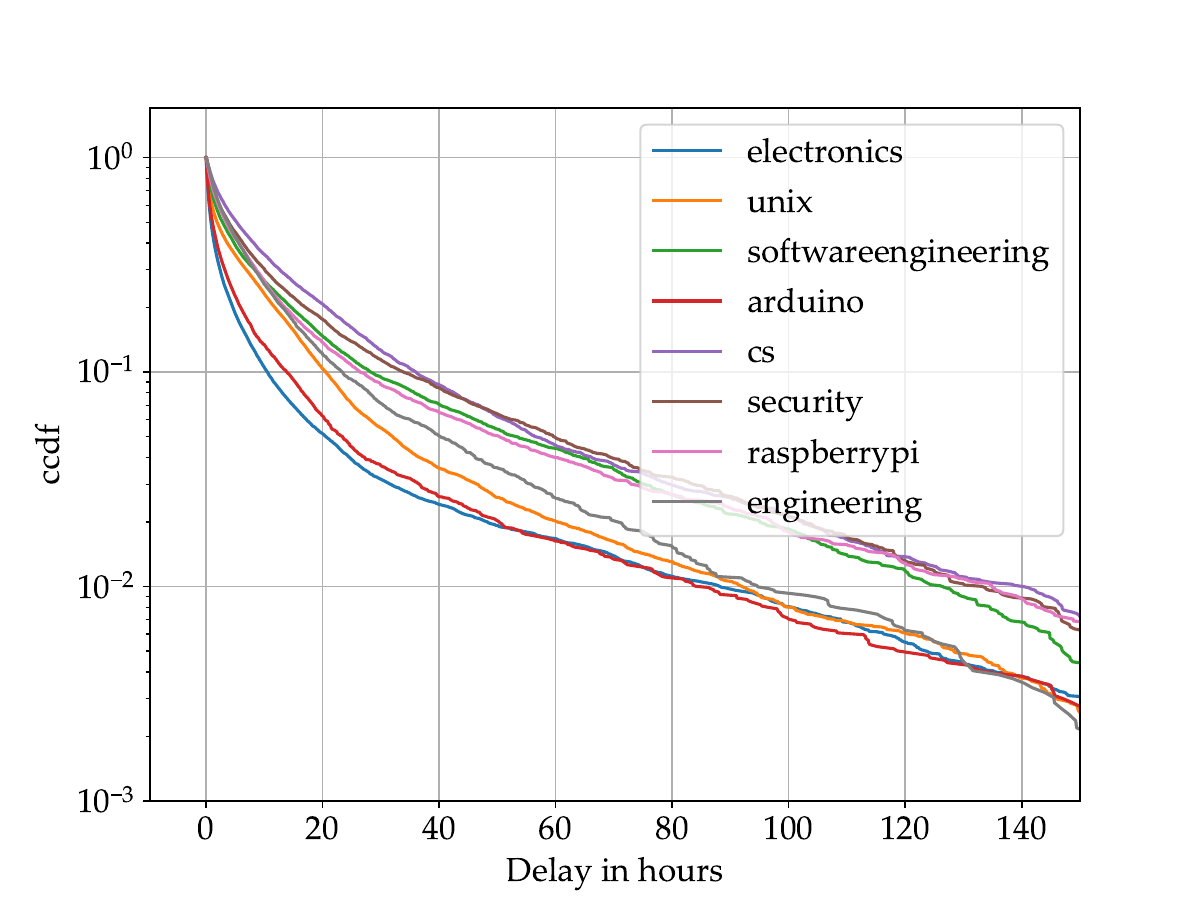}
\caption{Complementary cdf of Answering delays.}
\label{fig:answeringdelays}
\end{figure}

\section{Simulations}
\label{sec:simulation}
We obtained question-answering datasets from \cite{internetarchive} on 37 of the most active {\tt StackExchange} QA technical sites, each of which we consider a {\em topic} $x$ here. We used this dataset to calculate inter-arrival times between consecutive requests in each topic, for a total of 1.8 million such arrivals spanning more than 10 years for many of these sites. For each topic, we use these samples of inter-arrival times to simulate the request arrival process $a_x(t)$, noting that the distribution of these times may not be geometric. We also used the dataset to extract 200 most frequent answerers (`experts') in these sites in terms of number of questions answered, among those who are significantly active in at least two sites (such as `electronics' and `arduino'), in the sense of having at least 10\% of their answerers in each site. For these answerers, we obtained their answers and calculated the delay between the question being posed and answer being proposed, for a total of 270,000 such answers. For each answerer, we used her corresponding list of delays in different topics to simulate the time needed for her to answer that topic by drawing a sample at random with replacement from her list of delays, for each new request. Again,  we note that the resulting delay will not necessarily a geometric random variable. Figure \ref{fig:qix} shows the $q_i(x)$ functions calculated  for different experts, based on their average sample delays, with a few of the answerer's StockOverflow Account IDs marked on the figure. These answering rates are low because the time slot here is 8 minutes. The figure shows that a few answerers dominate the system, which seems unreasonable in our proposed knowledge worker-based system (since they may be paid), and will give uninteresting results here. So, we scaled the answering delay sample values of each expert by a different constant, so that each has the same sum answering rate $\sum_x q_i(x)$, while not changing their relative delays in different topics. Finally, it was observed that a certain fraction of answers come extremely late (often after several months), which may not be as interesting to users of the proposed system. So, we truncated the answering delays to a maximum of 96 hours - this affected 9\% of the answering delays. We observed that without such truncation (such as declaring some questions as too difficult), the system becomes unstable.
The Complementary cumulative distribution function (ccdf) of aggregate answering delays for some topics,  plotted in Figure \ref{fig:answeringdelays}, shows that while the distribution has an exponential distribution over a wide
range (thus, partially justifying our geometric random variable assumption), it does tends to cluster
near smaller values.

\begin{figure}[t]
\centering
\includegraphics[width=0.9\textwidth]{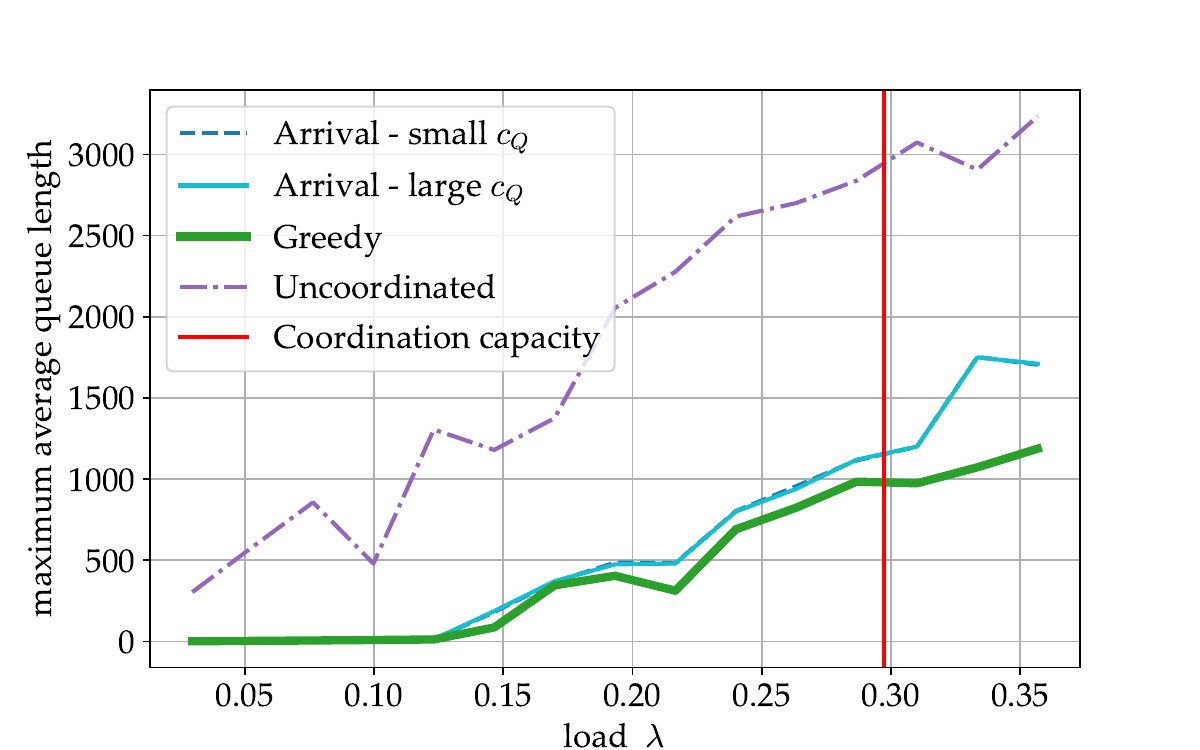}
\caption{Maximum of time-averaged topic queues versus $\lambda$.}
\label{fig:coordqueuelengths2}
\end{figure}

With such empirical request arrival processes and answering delays, we simulated the two federated schedulers for the Coordination case of Section \ref{sec:multipleonline}, where the load $\lambda$ is varied by scaling the  inter-arrival time variable generated using the dataset, as described above, and $n_i\equiv 1$. In our simulation, both schedulers, the Greedy and the Arrival-based one, work on the assigned request until completion, instead of returning it to the queues at the end of the slot. This is to allow us to check the sensitivity of our obtained capacity to this assumption. For comparison, the Coordination capacity \eqref{eqn:multiplecoordinating} was computed using $p(x)$ and $q_i(x)$ obtained from the average inter-arrival times in the different topics and the average delays for different topic-expert pairs.
Figure \ref{fig:coordqueuelengths2} shows the maximum  (over  topics) of time average of the queue lengths of the different schedulers using the empirical data, as outlined above, for different request loads $\lambda$, averaged over 10 separate trials of $10^5$ eight minute slots each. In all cases, the queue length is small, as long as the load is sufficiently below coordination capacity $\lambda^*$ (shown by vertical line). For the Arrival scheduler, the $c_Q$ value ($10^{-5}$ for `small' versus $1$ for `large') does not seem to affect the queues unduly.
For comparison, the Uncoordinated case, where the experts are allowed to randomly choose any waiting request in topic that they are historically significantly active in, performs worse than the Coordinated cases.

\begin{figure}[t]
\centering
\includegraphics[width=0.9\textwidth]{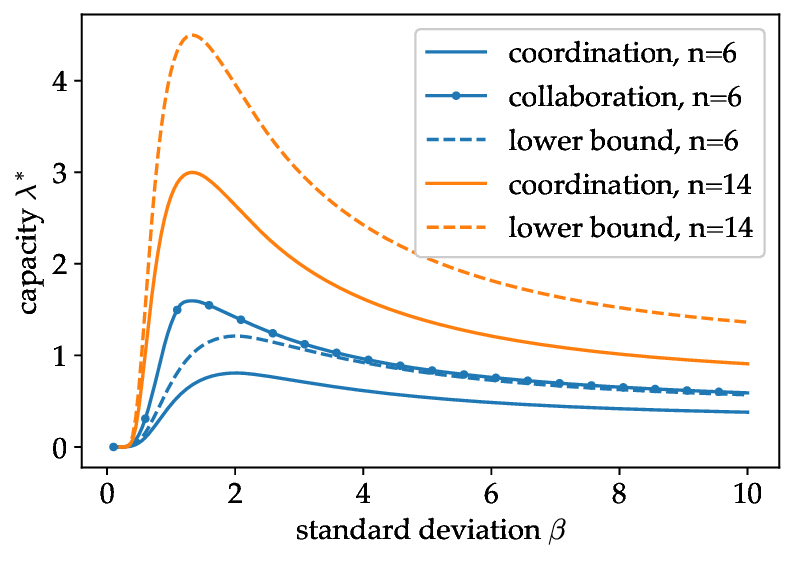}
\caption{Coordination and collaboration capacity versus $\beta$.}
\label{fig:coordcollabcap}
\end{figure}

Since we are not aware of any dataset as yet that can be used to infer collaboration expertise, we  numerically compute this capacity and contrast it with coordination capacity for a stylized model. We assume $|\xspace|=20$ topics, $n_i\equiv 1$ and $n=6$ versus $n=14$ experts, with the topics arranged in an interval, and with $q_i(x)$ modeled as a truncated Gaussian, normalized so that $\sum_x q_i(x)=1, \ \forall \,i$. The mode of the Gaussian for the experts is chosen uniformly spaced within the interval, to simulate the experts specialized in different topics. We assume all pairwise collaborations are allowed, and choose $q_S(x)=c \max_{i \in S} q_i(x)$ for pairwise $S$, with $c=3$. Figure \ref{fig:coordcollabcap} shows the coordination and collaboration capacities, as well as the lower bound on the collaboration capacity given by Lemma \ref{thm:collabcoordcomparison} part (c), as a function of the standard deviation $\beta$ of the truncated Gaussian. Since a smaller standard deviation results in experts having high answering rates, but for a narrower range of topics (i.e., highly specialized experts), the figure shows that both capacities are highest  when experts are neither too specialized, nor too broad. Moreover, the proposed lower bound is close to the exact collaboration capacity over a wide range of standard deviations.
With 14 experts, we could not compute the exact collaboration capacity numerically due to the exponentially large number of maximal independent sets, but the collaboration lower bound does show increased capacity compared to coordination (and also shows the value of that lower bound).

\section{Conclusions}
\label{sec:conclusions}

In this paper, we analyzed the request handling capacity of a QA forum, assuming that experts in the forum agree to use a scheduler to receive assignments of questions to research. We proposed two federated schedulers that achieve the coordination capacity, wherein experts coordinate on question assignment, but otherwise work separately. We also proposed a collaboration mode, and showed that with an appropriate scheduler, the capacity in this mode may be larger. Future work includes considering question-answering over social networks with arbitrary graph relationships between experts,  distributed scheduling  over such networks, and hybrid systems that allow experts some choice in selecting which questions to answer. 

\section{Acknowledgments}

This material is based upon work supported by the US National Science Foundation under Grant 1422193.

\vspace{6truemm}

\begin{thebibliography}{00}

\bibitem{cqa} I. Srba and M. Bielikova, ``A comprehensive survey and classification of approaches for Community Question Answering," {\em ACM Transactions on the Web}, pp. 1-63, August 2016.

\bibitem{socnetworkexamples} \url{ https://stackexchange.com/}, \url{ https://www.quora.com/}, \url{ https://answers.microsoft.com/}.

\bibitem{nassif} H. Nassif, M. Mohtarami and J. Glass, ``Learning semantic relatedness in Community Question Answering using neural models," {\em Proc. Workshop on Representation Learning for NLP}, pp. 137-147, August 2016.
\bibitem{dhamal} S. Dhamal, ``Effectiveness of Diffusing Information through a Social Network in Multiple Phases," {\em IEEE Globecom}, pp. 1-7, Dec. 2018.

\bibitem{fan} T. -H. Fan and K. -C. Chen, ``A New Social Network Model of Online Forums," {\em IEEE Globecom}, pp. 1-6, Dec. 2017.

\bibitem{internetarchive} {\tt  https://archive.org/download/stackexchange/}.


\bibitem{nakov} P. Nakov, et. al., ``SemEval-2015 Task 3: Answer Selection in Community Question Answering," {\em Wksp. on Semantic Evaluation}, June 2015.

\bibitem{huang} C. Huang, H. Yu, J. Huang and R.  Berry, ``Crowdsourcing with Heterogeneous Workers in Soc. Net.s," {\em IEEE Globecom}, pp. 1-6, Dec. 2019.


\bibitem{negi2022} R. Negi and M. Yilmaz, ``A Theoretical Framework for Information Search using Online Social Networks,'' {\em IEEE Globecom - Social Networks}, pp. 6421-6426, Dec. 2022.

\bibitem{meyn} S. Meyn,  Control Techniques for Complex Networks, Cambridge University Press, 2007.

\bibitem{han} F. Han, et. al., ``Distributed representations of expertise,” {\em Proc. SIAM Int. Conf. on Data Mining}, pp. 531-539, June 2016.

\bibitem{bremaud} P. Bremaud, Markov chains: Gibbs fields, Monte Carlo simulation, and queues, Springer, New York, 1999.



\bibitem{hap} R. Borndörfer, O. Heismann, ``The hypergraph assignment problem," {\em Discrete Optimization},
    vol. 15, pp. 15-25, Feb. 2015.



\end{thebibliography}
\end{document}